\documentclass[sigconf]{acmart}

\usepackage{balance}

\def\BibTeX{{\rm B\kern-.05em{\sc i\kern-.025em b}\kern-.08emT\kern-.1667em\lower.7ex\hbox{E}\kern-.125emX}}
    
\copyrightyear{2019} 
\acmYear{2019} 
\acmConference[MM '19]{Proceedings of the 27th ACM International Conference on Multimedia}{October 21--25, 2019}{Nice, France}
\acmBooktitle{Proceedings of the 27th ACM International Conference on Multimedia (MM '19), October 21--25, 2019, Nice, France}
\acmPrice{15.00}
\acmDOI{10.1145/3343031.3350875}
\acmISBN{978-1-4503-6889-6/19/10}

\usepackage{epsfig}
\usepackage{graphicx}
\usepackage{amsmath}
\usepackage{amssymb}

\usepackage{url}

\makeatletter
\def\url@leostyle{%
	\@ifundefined{selectfont}{\def\UrlFont{\sf}}{\def\UrlFont{\small\ttfamily}}}
\makeatother

\usepackage{subfigure}
\usepackage{multirow}
\usepackage{caption}
\usepackage{booktabs}

\newcommand\ies{\textit{i.e.}}
\newcommand\egs{\textit{e.g.}}
\newcommand\etals{\textit{et al.}}

\newcommand\figcaption{\def\@captype{figure}\caption}
\newcommand\tabcaption{\def\@captype{table}\caption}

\begin{document}

\fancyhead{}

\title{Matching Images and Text with Multi-modal Tensor Fusion and Re-ranking}

\author{Tan Wang}
\affiliation{%
  \institution{Center for Future Media and School of Information and Communication Engineering \\
  	University of Electronic Science and Technology of China, China}
}

\author{Xing Xu}
\authornote{Corresponding author.}
\affiliation{%
 \institution{Center for Future Media and School of Computer Science and Engineering 
 	\\ University of Electronic Science and Technology of China, China}
}

\author{Yang Yang}
\affiliation{%
	\institution{Center for Future Media and School of Computer Science and Engineering 
		\\ University of Electronic Science and Technology of China, China}
}

\author{Alan Hanjalic}
\affiliation{%
	\institution{Multimedia Computing Group \\ Delft University of Technology, The Netherlands}
}

\author{Heng Tao Shen}
\affiliation{%
	\institution{Center for Future Media and School of Computer Science and Engineering 
		\\ University of Electronic Science and Technology of China, China}
}

\author{Jingkuan Song}
\affiliation{%
	\institution{Center for Future Media and School of Computer Science and Engineering 
		\\ University of Electronic Science and Technology of China, China}
}
 
%
%
%

%
\renewcommand{\shortauthors}{Tan Wang and Xing Xu, et al.}

%
\begin{abstract}
A major challenge in matching images and text is that they have intrinsically different data distributions and feature representations. 
Most existing approaches are based either on embedding or classification, the first one mapping image and text instances into a common embedding space for distance measuring, and the second one regarding image-text matching as a binary classification problem. 
Neither of these approaches can, however, balance the matching accuracy and model complexity well. 
We propose a novel framework that achieves remarkable matching performance with acceptable model complexity.
Specifically, in the training stage, we propose a novel Multi-modal Tensor Fusion Network (MTFN) to explicitly learn an accurate image-text similarity function with rank-based tensor fusion rather than seeking a common embedding space for each image-text instance. 
Then, during testing, we deploy a generic Cross-modal Re-ranking (RR) scheme for refinement without requiring additional training procedure.
Extensive experiments on two datasets demonstrate that our MTFN-RR consistently achieves the state-of-the-art matching performance with much less time complexity.
\end{abstract}

%
%


%

%

%
\maketitle

\begin{figure}[htb]
	\centering
	\includegraphics[width=0.43\textwidth]{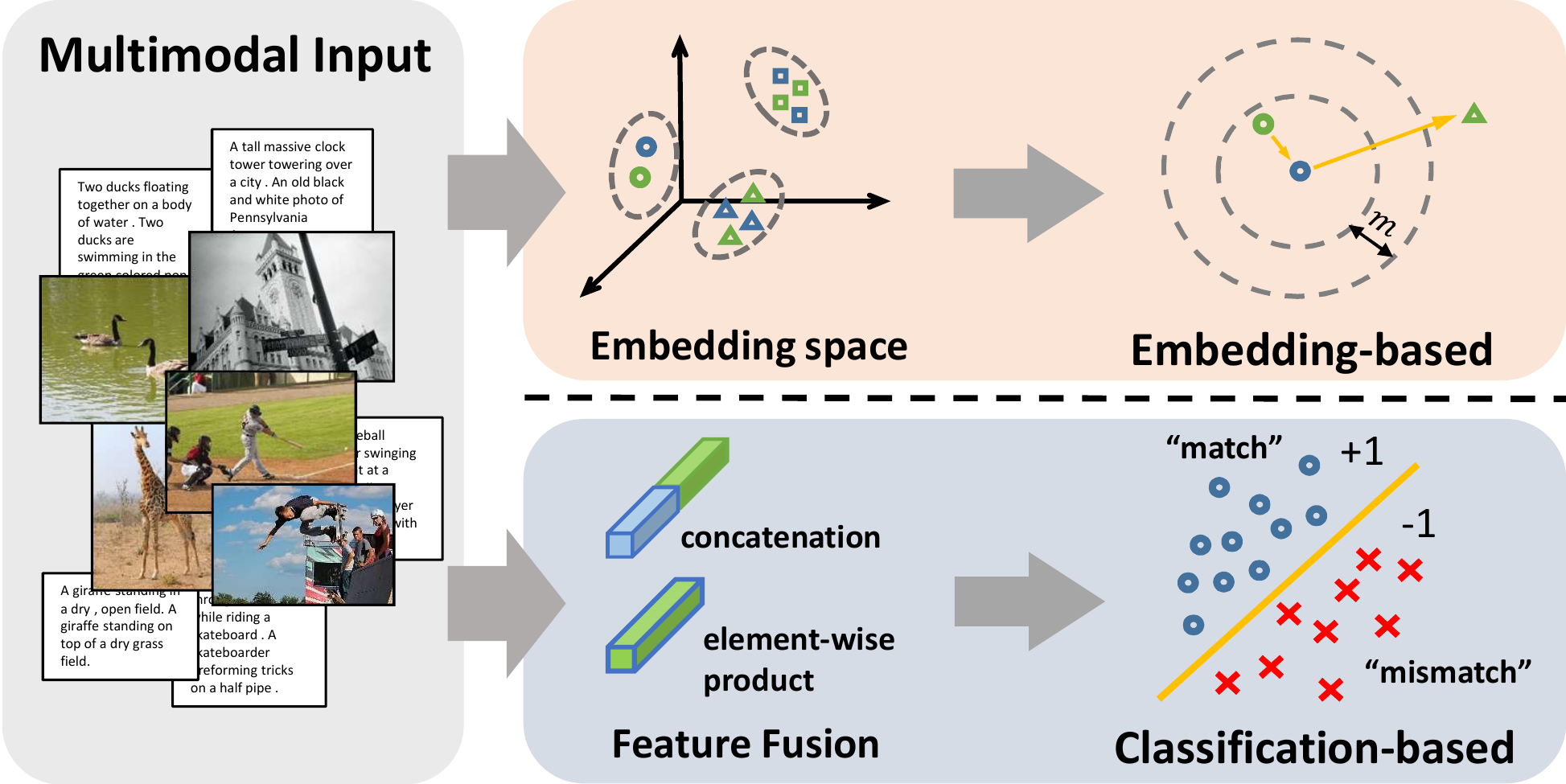}
	\caption{Illustration of the embedding-based (top) and the classification-based methods (bottom) that are typically used for image-text matching.}	
	\label{fig:framework_intro1}
\end{figure}

\section{Introduction}
\label{sec:intro}
In contrast to retrieval of unimodal data, image-text matching \cite{CMPM_ECCV2018, IALC_ICCV2017, sm-LSTM_CVPR2017,SCAN_ECCV2018} focuses on retrieving the relevant instances of a different media type than the query, including two typical \textit{cross-modal retrieval} \cite{song_imh2013,xu_wwwj2018,hu2019collective,xu2019tcyb} scenarios: 1) image-based sentence retrieval (I2T), \ies, retrieving ground-truth text sentences given a query image, and 2) text-based image retrieval (T2I), \ies, retrieving matched images given a query text.
Essentially, image-text matching requires algorithms that are able to assess the similarity between data and feature representations of images and and the semantics of text. 
Due to large discrepancy between the nature of textual and visual data and their feature representations, achieving this matching in an effective, efficient and robust fashion is a challenge. 

A straightforward step in pursuing this challenge is to expand a typical unimodal \textit{classification approach} to operate in a cross-modal case. Methods like \cite{CMPM_ECCV2018, DSPE_CVPR2016, IALC_ICCV2017, sm-LSTM_CVPR2017} have been proposed to predict match or mismatch (\ies, ``+1'' and ``-1'') for an image-text pair input by optimizing a logistic regression loss, turning this into a binary classification problem. They have been, however, shown to be insufficiently capable of handling cross-modal data complexity and therefore insufficiently effective in finding boundaries between unbalanced matching and non-matching image-text pairs. As an alternative, \textit{embedding-based approach} has therefore been investigated as well. Embedding-based methods (\egs, \cite{DVSA_CVPR2015,SCAN_ECCV2018,DPC_arxiv2017,vse++, SCO_CVPR2018}) try to map image and text features, either global or local, into a joint embedding subspace by optimizing a ranking loss that ensures the similarities of the ground-truth image-text pairs to be greater than that of any other negative pairs by a margin $m$. Once the common space is established, the relevance between any image and text instance can be easily measured by cosine similarity or Euclidean distance. However, the main limitation of these methods is that constructing such high-dimensional common space for the complex multi-modal data is not a trivial task, typically showing significant problems with learning convergence and generalizability of the learned space and requiring significant computational time and resources. The general pipelines of the two categories of approaches are illustrated in Figure \ref{fig:framework_intro1}. 

\begin{figure*}[!htb]
	\centering
	\includegraphics[width=0.9\textwidth]{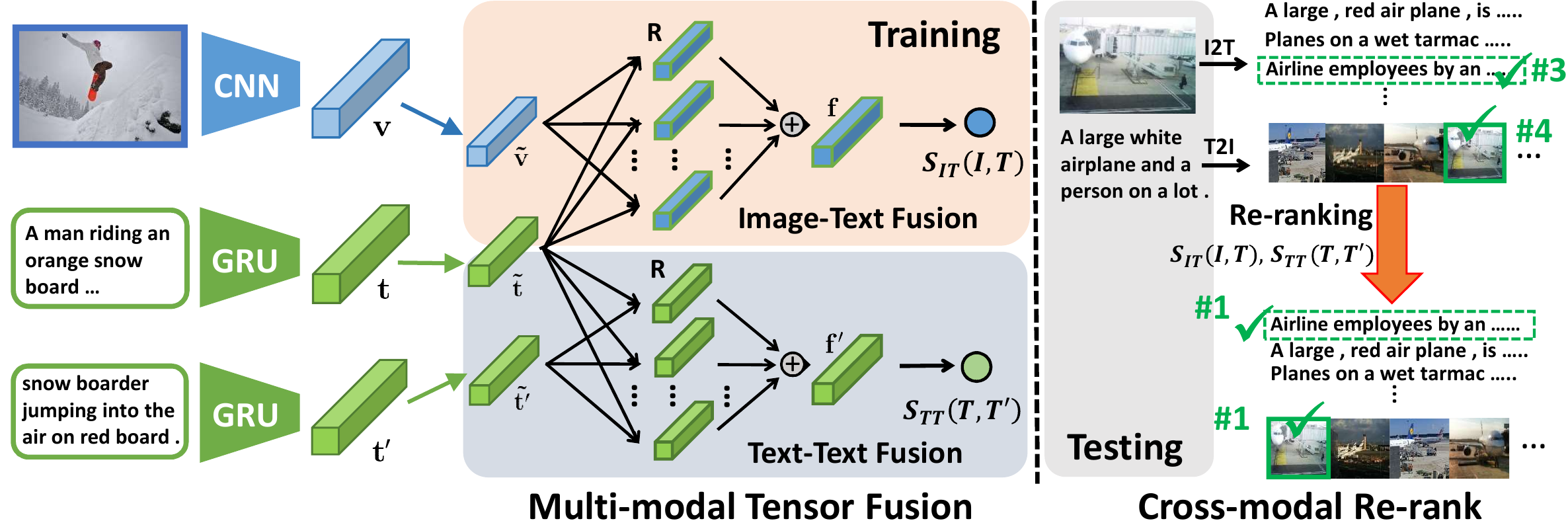}
	\caption{The overview architecture of our proposed framework separated by training and testing parts. 1) During training, the global features of multi-modal inputs are firstly passed into two branches (\ies, \textit{Image-Text Fusion} and \textit{Text-Text Fusion}). Then for each branch a tensor fusion scheme with rank constraint is used for modelling rich interactions between the input features to a vector for learning the similarity score. 2) In testing, a cross-modal re-ranking scheme is applied to jointly consider the I2T and T2I retrieval, with the combination of both image-text similarity $S_{IT}\left(I,T\right)$ and text-text similarity $S_{TT}\left(T,T'\right)$.}
	\label{fig:framework_all}
\end{figure*}

Clearly, while the embedding-based methods have more potential to capture the complexity in data than the classification-based methods, their model and algorithmic complexity is significantly higher. This analysis leads to a question that inspired the research reported in this paper: \textit{Is a new image-text matching framework possible that combines the advantages of the two paradigms, \ies, balancing the matching performance and model efficiency?}
To answer this question, in this paper we propose a novel image-text matching framework named \textit{Multi-modal Tensor Fusion Network with Re-ranking} (MTFN-RR) that in an innovative fashion combines the concepts of embedding and classification to achieve the aforementioned balance. As illustrated in Figure \ref{fig:framework_all}, our framework is constructed as a cascade of two steps: 1) deploying tensor fusion to learn an accurate image-text similarity measure in the training stage, and 2) performing cross-modal re-ranking to jointly refine the I2T and T2I retrieval results in the testing stage.

For the first part, our MTFN takes the multi-modal global feature as input and then passes them to two branches of \textit{Image-Text Fusion} and \textit{Text-Text Fusion}.
Then for each branch, a tensor fusion block with rank constraint is used to capture rich bilinear interactions between multimodal input into a vector.
Finally, the similarity of input is directly learned with a fully convolutional layer from the fused vector and naturally embedded to the advanced ranking loss to encourage the large-margin between groundtruth image-text pairs and negative pairs.
In this way, the similarity measuring functions from both image-text and text-text input are directly learned without constructing the whole embedding subspace.

Regarding the re-ranking step in the second part, we note that in the previous work, the I2T and T2I retrieval tasks are typically conducted separately in the testing phase.
This may be problematic because in the training stage these two tasks are optimized with a bi-directional max-margin ranking loss function like Eq. \ref{eq:loss_i2t}, yielding a discrepancy between training and inference.
To reduce this discrepancy in an efficient way, we develop an general cross-modal re-ranking (RR) scheme that jointly considers the retrieval results of both I2T and T2I directions, to bridge the gap between training and testing with little time but achieving significant improvement applicable to most existing image-text matching approaches.
Additionally, to mitigate the effects of the unbalanced data (\ies, an image corresponds to five sentences) in MSCOCO \cite{MSCOCO} and Flickr30k \cite{flickr30k}, the similarities between unimodal text predicted by our MTFN are further used to significantly boost the T2I retrieval performance.

We summarize our contributions as follows: 1) We propose a novel Multi-modal Tensor Fusion Network (MTFN) that directly learns an accurate image-text similarity function for visual and textual global features via image-text fusion and text-text fusion. It explicitly incorporates the advantages of both embedding-based and classification-based methods and enables efficient training. 
2) We further develop an efficient cross-modal re-ranking scheme that remarkably improves the matching performance without extra training  and can be freely applied to other off-the-shelf methods.
With extensive experiments, the proposed MTFN itself shows competitive accuracy compared to the current best methods on two standard datasets 
with much less time complexity and simpler feature extraction.
Furthermore, when integrating the proposed RR scheme in our MTFN, it achieves the state-of-the-art performance on two datasets, especially on the R@1 score, showing the effectiveness of the RR scheme.
The implementation code and related materials are available at \url{https://github.com/Wangt-CN/MTFN-RR-PyTorch-Code}.

\section{Related Work}
\noindent\textbf{Image-Text Similarity.}
As mentioned before,
the embedding-based methods
\cite{DVSA_CVPR2015,SCAN_ECCV2018,hm-lstm_iccv2017}
project multimodal (global/local) features into a common embedding space, in which similarities between instances are measured by conventional cosine or Euclidean distance.
However, modeling the similarity between image and text can also be regarded as a classification problem to directly answer whether two input samples match each other \cite{two_branch_TPAMI2018, MCB_EMNLP2016, vqa_class1}.
These methods typically secure rapid convergence of the training process, but are limited to fully exploit the identity information of cross-modal features with simple match/mismatch classifiers.
Our MTFN is proposed to leverage the advantages of the two kinds of methods above.
Specifically, a fully fusion network is firstly designed for directly learning a similarity function from the image-image fusion and text-text fusion rather than using conventional distance metric in a common embedding space.
The predicted similarity is equipped to the widely used ranking loss with large margin constraint, which can be optimized efficiently in the next step.

\noindent\textbf{Multi-modal Fusion.}
To fully capture the interactions between multiple modalities, a number of fusion strategies have been used for exploring the relationship of visual and textual data.
Liu \etals~\cite{RRF-Net_ICCV2017} applied a fusion module to integrate the intermediate recurrent outputs and generate a more powerful representation for image-text matching.
Wang \etals~\cite{two_branch_TPAMI2018} used element-wise product to aggregate features from visual and text data in two branches.
Recently, bilinear fusion \cite{MCB_EMNLP2016, MLB_ICLR2017, MUTAN_ICCV2017}
has proved to be more effective than traditional fusion scheme such as element-wise product \cite{vqa_fusion1}
and concatenation \cite{vqa_fusion2} in Visual Question Answering (VQA) problem, since it enables all elements of multi-modal vectors to interact with each other in a multiplicative way.
We draw inspiration from the VQA method \cite{MUTAN_ICCV2017} to capture the bilinear interactions of the image-text and text-text data inputs and directly learn image-text similarity.
Note that instead of modelling a vector between two modalities by tucker decomposition for classification with a small set of concepts/answers in VQA dataset, here our MFTN can be regarded as a general tensor fusion architecture for various inputs (\egs, image-text, text-text) to directly learn the similarity.

\noindent\textbf{Re-ranking Scheme.}
Re-ranking has been successfully studied in unimodal retrieval task such as person re-ID \cite{person_rerank1, person_rerank2, person_rerank3, person_rerank4}, object retrieval \cite{object_ret1, object_ret2} and text-based image search \cite{yang2010supervised, yang2012prototype, hsu2006video} to improve the accuracy.
In these problems, retrieved candidates within initial rank list can be re-ordered  as an additional refinement process.
For example, Leng \etals~\cite{person_rerank4} proposed a bi-directional ranking method with the newly computed similarity by fusing the contextual similarity between query images.
Zhong \etals~\cite{k-encoding_cvpr2017} designed a new feature vector for the given image under the Jaccard distance after the initial ranking.
Yang \etals~\cite{yang2010supervised} introduced a supervised ``learning to rerank'' paradigm into the visual search reranking learning by applying query-dependent reranking features.  
Unlike the unimodal re-rank methods and learning to rerank paradigm in text based image search, we propose a cross-modal re-ranking scheme for image-text matching scenario without supervision and learning procedure, which combines the bidirectional retrieval process (I2T and T2I), only takes few seconds and can be inserted in most image-text matching methods for performance improvement.

\section{Proposed Model}
Let $\mathcal{O} = \{(I_n, T_n)\}_{n = 1}^N$ be a training set of $N$ image-text pairs, where the image set is denoted as $\mathcal{X} = \{I_n\}_{n = 1}^N$ and the text set as $\mathcal{Y} = \{T_n\}_{n = 1}^N$.
We refer to $(I_p, T_p)$ as positive pairs and $(I_p, T_{q \neq p})$ as negative pairs, \ies, the most relevant sentence to image $I_p$ is $T_p$ and for sentence $T_p$, its matched image is $I_p$.
Given a query of one modality, the goal of image-text matching is to find the most relevant instances of the other modality.
In this work, we define a similarity function $S(I, T) \in \mathbb{R}^1$ that is expected to, ideally, assign higher similarity scores to the positive pairs than the negative ones.
This procedure can be derived as:
\begin{equation}
\label{eq:cross-model retrieval task}
S(I_{p}, T_{p}) > S(I_{p}, T_{q}), ~~ \forall I_p \in \mathcal{X}; ~~\forall T_p, T_q \in \mathcal{Y}.
\end{equation}
Accordingly, we can conduct I2T retrieval task by ranking a database of text instances based on their similarity scores with the query image using $S(I, T)$, and likewise for T2I retrieval task.
Different from most existing embedding-based methods that adopt conventional distance metric (\egs, cosine similarity or Euclidean distance) as the similarity function on a common embedding space, in this work, we aim to directly learn a similarity function that accurately measures the relevance of image-text pairs without seeking for the common subspace for each instance.

\subsection{Multi-modal Tensor Fusion Network}
Inspired by the Multimodal Tucker Fusion proposed in visual question answering \cite{MUTAN_ICCV2017}, as illustrated in Fig. \ref{fig:framework_all} we introduce a novel Tensor Fusion Network into image-text matching for feature merging and similarity learning.    
Specifically, MTFN contains two branches of \textit{Image-Text Fusion} and \textit{Text-Text Fusion}, learning the similarity scores with different inputs (\ies, image-text and text-text). 
The Image-Text Fusion branch is a conventional process that fuses the global feature representations of images and sentences on dimension of tensor and predicts the similarity score of any image-text pair as $S_{IT}(I, T)$.
Moreover, considering the fact that multiple sentences annotated to one image have common semantics, we introduce the Text-Text Fusion branch to further capture the semantic relevance of any text-text pairs as $S_{TT}(T, T')$.
Different with \cite{two_branch_TPAMI2018}, the learned information of text modality would be used for re-ranking in testing stage for narrowing the gap between training and inference.
In the following parts, we will explicitly depict the details of the two fusion branches in our MTFN.

\noindent\textbf{Image-Text Fusion.}
Firstly, given the global feature vectors $\mathbf{v}$ and $\mathbf{t}$ for image $I_p$ and sentence $T_q$, the intra-modal projection matrices $\boldsymbol{\mathrm{W}}_{v}$ and $\boldsymbol{\mathrm{W}}_{t}$ are constructed to encode two feature vectors into spaces of respective dimensions $d_{v}$ and $d_{t}$, which can be written as $\mathbf{\tilde{v}} = \boldsymbol{\mathrm{W}}_{v} \mathbf{v} \in \mathbb{R}^{d_{v}}$ and $\mathbf{\tilde{t}} = \boldsymbol{\mathrm{W}}_{t} \mathbf{t} \in \mathbb{R}^{d_{t}}$.
Then for feature fusion at the tensor level, we project $\mathbf{\tilde{v}}$ and $\mathbf{\tilde{t}}$ into a common space and merge them with an element-wise product, which can be written as:
\begin{equation}
\label{eq:elementwise}
\mathbf{f} = \left( \boldsymbol{\mathrm{W}}_{\tilde{v}} \mathbf{\tilde{v}} \right) \odot \left( \boldsymbol{\mathrm{W}}_{\tilde{t}} \mathbf{\tilde{t}} \right),
\end{equation}
where $\boldsymbol{\mathrm{W}}_{\tilde{v}} \in \mathbb{R}^{d_{v} \times d_{f}}$, $\boldsymbol{\mathrm{W}}_{\tilde{t}} \in \mathbb{R}^{d_{t} \times d_{f}}$
and $\odot$ denotes the element-wise product in matrices.

Actually, considering that each fusion vector $\mathbf{f} \in \mathbb{R}^{d_{f}}$ can be regarded as a rank-1 vector carrying limited information, to fully encode the bilinear interactions between the two modalities, we further impose a rank constraint $R$ on the fusion vector $\mathbf{f}$ to express it as a sum of $R$ rank-1 vectors instead of performing a single feature merging function.
In this way, we directly learn $R$ different common subspaces.
For each space embedding $\boldsymbol{\mathrm{W}}^{r}_{\tilde{v}}$ and $\boldsymbol{\mathrm{W}}^{r}_{\tilde{t}}, r \in \left(1, R\right)$, a specific fusion vector can be obtained by element-wise product and ultimately summed together, allowing the model to jointly capture the interactions between two modalities from different representation subspaces.
Thus the Eq. \ref{eq:elementwise} can be rewritten as follows:
\begin{equation}
\label{elementwise_r}
\mathbf{f} = \sum_{r=1}^{R} \left( \boldsymbol{\mathrm{W}}^{r}_{\tilde{v}} \mathbf{\tilde{v}} \right) \odot \left( \boldsymbol{\mathrm{W}}^{r}_{\tilde{t}} \mathbf{\tilde{t}} \right).
\end{equation}
Finally a fully connected layer $\boldsymbol{\mathrm{W}}_{m}$ is added to transform the fusion vector $\mathbf{f}$ to the score $S_{IT}$ which infers the similarity between image and text followed by a sigmoid layer embedding to $\left(0, 1\right)$:
\begin{equation}
\label{sigmoid}
S_{IT}(I_p, T_q) =Sigmoid\left(\boldsymbol{\mathrm{W}}_{m} \mathbf{f}\right).
\end{equation}

Next, instead of treating the ``match'' and ``mismatch'' as a binary classification problem, we propose to naturally combine the similarity $S_{IT}\left(I,T\right)$ between the two inputs with the widely used ranking loss constraint in existing embedding-based methods to construct a bi-directional max-margin ranking loss.
By this way the nonlinear boundary can be easier found while ensuring the preservation of inter-modal invariance simultaneously, which will be further explained in \textit{Ablation Study}.
Specifically, in our work, we follow \cite{vse++, SCAN_ECCV2018} to focus on the hardest negatives in the mini-batch during training.
For each positive pair of an image and a text $\left(I_{p}, T_{p}\right)$, we additionally sample their hardest negatives which are given by $I_{h} = \arg\max_{h \neq p} S_{IT} \left(I_{h}, T_{p}\right)$ and $T_{h} = \arg\max_{h \neq p} S_{IT} \left(I_{p}, T_{h}\right)$.
Then the image-text loss can be defined as:
\begin{equation}
\label{eq:loss_i2t}
\begin{aligned}
L\left(I_{p}, T_{p}\right) & = [\alpha - S_{IT}\left(I_{p}, T_{p}\right) + S_{IT}\left(I_{p}, T_{h}\right)]_{+} \\
& + [\alpha - S_{IT}\left(I_{p}, T_{p}\right) + S_{IT}\left(I_{h}, T_{p}\right)]_{+},
\end{aligned}
\end{equation}
where $\alpha$ is a constant value of the margin and the operator $[z]_{+} = \max{\left(0, z\right)}$ compares the tolerance value with zero.
By minimizing the loss term in Eq. \ref{eq:loss_i2t}, the network is trained to guarantee that the truly matching image-text pairs have larger similarity scores than the most confusing negative pairs by a margin $\alpha$.

\noindent\textbf{Text-Text Fusion.}
Different from the \textit{Image-Text Fusion}, the \textit{Text-Text Fusion} measures the similarity of two unimodal sentences, \ies, $T$ and $T'$, whose features are respectively denoted as $\mathbf{t}$ and $\mathbf{t'}$.
Similar to the tensor fusion in Eq. \ref{elementwise_r}, here the similarity function of two sentences can be derived as:
\begin{equation}
\label{eq:sim_t2t}
S_{TT}\left(T, T'\right) = Sigmoid\left(\boldsymbol{\mathrm{W}}_{m'} \sum_{r=1}^{R} \left( \boldsymbol{\mathrm{W}}^{r}_{\tilde{t}} \mathbf{\tilde{t}} \right) \odot \left( \boldsymbol{\mathrm{W}}^{r}_{\tilde{t'}} \mathbf{\tilde{t'}} \right)\right),
\end{equation}
where $\mathbf{\tilde{t}}, \mathbf{\tilde{t'}} \in \mathbb{R}^{d_{t}}$ and $\boldsymbol{\mathrm{W}}^{r}_{\tilde{t}}, \boldsymbol{\mathrm{W}}^{r}_{\tilde{t'}} \in \mathbb{R}^{d_{t} \times d_{f'}}$.
Accordingly, we also adopt the ranking constraint with large-margin to learn the text-text similarity $S_{TT}\left(T, T'\right)$.
Specifically, given two sentences in a  positive pair $\left(T_{p}, T_{q}\right)$, they have the same negative sample $T_{h}$.
Like Eq. \ref{eq:loss_i2t},  here the text-text loss can be formulated as:
\begin{equation}
\label{eq:loss_t2i}
L\left(T_{p}, T_{q}\right) = [\alpha - S_{TT}\left(T_{p}, T_{q}\right) + S_{TT}\left(T_{p}, T_{h}\right)]_{+},
\end{equation}
which becomes a triplet ranking loss term.
The two loss functions in Eq. \ref{eq:loss_i2t} and Eq. \ref{eq:loss_t2i} can be optimized independently with Adam optimizer \cite{adam}.


\subsection{Cross-modal Re-ranking}
Like most existing methods, the I2T and T2I retrieval tasks can be conducted in our MTFN separately using the learned similarity function $S_{IT}(I, T)$ to obtain the retrieval candidates for an query image or text.
However, the interactions between bi-directional retrieval (I2T and T2I) are ignored in the testing stage, resulting in a discrepancy between training and inference.
Motivated by the success of deploying RR methods in person Re-id task \cite{k-encoding_cvpr2017, person_rerank1, object_ret2} and text-based image search \cite{yang2010supervised, yang2012prototype, hsu2006video} that are designed for retrieval within unimodal data, here we propose a cross-modal RR formulated as a novel $k$-reciprocal nearest neighbours searching problem to make the best of the initial learned similarity of image-text pairs and text-text pairs $S_{IT}\left(I, T\right), S_{TT}\left(T, T'\right)$ and narrow the gap between training and testing.

The basic assumption is that if an image is paired with a text, they can be retrieved from each other by I2T or T2I retrieval forwardly and reversely.
In other words,  for an image, its matching text should be the top of its ranking candidates and vice versa.
Based on this assumption, we define two re-ranking strategies of \textit{I2T Re-ranking} and \textit{T2I Re-ranking} as follows.
\begin{figure}[!htb]
	\centering
	\includegraphics[width=0.45\textwidth]{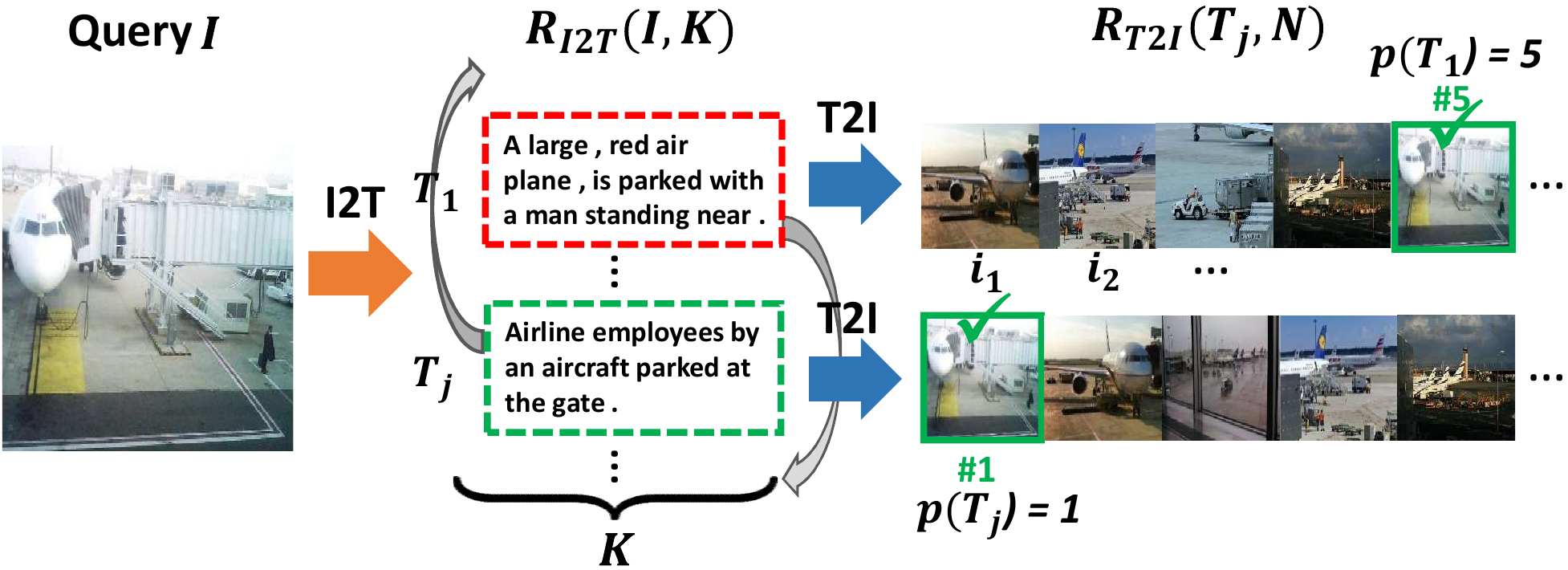}
	\caption{An example of our I2T Re-ranking scheme. Given a query image $I$, a conventional I2T retrieval list $R_{I2T}(I, K)$ is built firstly. Then we apply the inverse retrieval direction T2I for further refinement.}	
	\label{fig:proposal}
\end{figure}

\noindent\textbf{I2T Re-ranking.}
As shown in Fig. \ref{fig:proposal}, given a query image $I$ and its initial ranking list produced by our MTFN using $S_{IT}(I,T)$, we define $R_{I2T}\left(I, K\right)$ as the initial cross-modal $K$-nearest neighbour text of image $I$:
\begin{equation}
R_{I2T}(I, K) = \{T_1, ..., T_j, ..., T_K\},
\end{equation}
where $|R_{I2T}(I, K)| = K$ denotes the number of candidates in the list.
Then for each candidate text $T_{j}$, a set $R_{T2I}\left(T_j, N\right)$ of $N$-nearest images can be defined as:
\begin{equation}
R_{T2I}\left(T_j, N\right) = \{I_1, ..., I_k,...,I_N\},
\end{equation}
where $N$ is the number of images in testing set.
To fuse the bi-directional nearest neighbours of $R_{I2T}$ and $R_{T2I}$, we further introduce a position index for each candidate $T_j$ as:
\begin{equation}
p\left(T_j\right) = k,~ ~if~~ I_{k} = I,~ ~I_k \in R_{T2I}\left(T_j, N\right).
\end{equation}
Then a position set $P$ can be built for all candidate text in the initial $k$-nearest neighbours $R_{I2T}\left(I, K\right)$:
\begin{equation}
P\left(I,K\right) = \{p(T_1), ..., p(T_j), ..., p(T_K) \}.
\end{equation}
The set $P\left(I, K\right)$ can be regarded as a reordering of the initial retrieval list $R_{I2T}\left(T_j, N\right)$ using text modality, deploying the learned similarity matrix from the other direction (\ies, $T2I$) effectively. 
Therefore, we then just re-calculate the pairwise similarity between the query image $I$ and candidate text $T_{j}$ by ranking the position set $P$ as:
\begin{equation}
R_{I2T}^{'} = ranking\left(P\left(I,K\right)\right),
\end{equation}
where $R_{I2T}^{'}$ denotes the refined retrieval list for the query image $I$ after I2T re-ranking.

\noindent\textbf{T2I Re-ranking.}
As an image is commonly annotated with multiple sentences in datasets, we apply the obtained unimodal text similarity $S_{TT}(T,T')$ as a prior information to refine the T2I Re-ranking process.
Likewise, we first define the $k$-nearest images for a query text $T$ with initial ranking list generated by our MTFN using $S_{IT}(I,T)$:
\begin{equation}
R_{T2I}(T, K) = \{I_1, ..., I_j, ..., I_K\}.
\end{equation}
Differently, since considering that each image is annotated with multiple sentences in practice, the retrieval results of $T$ would have inner associations to those of other semantically related text.
Therefore, we find the unimodal nearest neighbour set $G\left(T, K'\right)$ of $T$ to replace the individual query text $T$ using the text-text similarity $S_{TT}\left(T, T'\right)$, as
\begin{equation}
G\left(T, K'\right) = \{T_1,T_2, ..., T_{K'}\},
\end{equation}
where $K'$ is the number of related text to $T$.
Then similar as the re-ranking procedure in \textit{I2T Re-ranking}, the refined  results are obtained by performing I2T retrieval for each image in $R_{T2I}(T, K)$, where the detailed steps are depicted as follows:
\begin{align}
\label{T2I_rerank}
&R_{I2T}\left(I_j, N\right) = \{T_1, ..., T_k, ..., T_N\} \nonumber\\
&p\left(I_j\right) = k,~ ~if~ T \in G\left(T_k, K'\right),~ ~T_k \in R_{I2T}\left(I_j, N\right)  \nonumber\\
&P\left(T,K\right) = \{p\left(I_1\right), ..., p\left(I_j\right),...,p\left(I_K\right) \}  \nonumber\\
&R_{T2I}^{'} = ranking\left(P\left(T,K\right)\right),
\end{align}
where $R_{T2I}^{'}$ is the refined image list for the query text $T$  after the T2I re-ranking.

\begin{table*}[!htb]
	\small
	\caption{Overall comparison with the state-of-the-art results. Three panels are the classification-based methods, embedding-based methods and our proposed method, respectively. The best results are marked in bold font.}
	\begin{tabular}{l|ccc|ccc|c|ccc|ccc|c}
		\hline \hline
		\multirow{3}{*}{Method} & \multicolumn{7}{c|}{Flickr30k dataset}                                                                        & \multicolumn{7}{c}{MSCOCO dataset}                                                                        \\ \cline{2-15}
		& \multicolumn{3}{c|}{I2T} & \multicolumn{3}{c|}{T2I}          & \multirow{2}{*}{mR} & \multicolumn{3}{c|}{I2T}       & \multicolumn{3}{c|}{T2I} & \multirow{2}{*}{mR} \\ \cline{2-7} \cline{9-14}
		& R@1         & R@5         & R@10        & R@1           & R@5           & R@10          &                     & R@1           & R@5           & R@10          & R@1           & R@5           & R@10 &                     \\ \hline
		LTBN (Sim) \cite{two_branch_TPAMI2018} (TPAMI'18)     & 16.6        & 38.8        & 51.0        & 7.4          & 23.5          & 33.3          & 28.4                & 30.9          & 61.1          & 76.2          & 14.0          & 30.0          & 37.8 	& 41.7                \\
		sm-LSTM \cite{sm-LSTM_CVPR2017} (CVPR'17)               & 42.5        & 71.9        & 81.5        & 30.2          & 60.4          & 72.3          & 59.8                & 53.2          & 83.1          & 91.5          & 40.7          & 75.8          & 87.4 & 72.0                \\ 
		CMPM \cite{CMPM_ECCV2018} (ECCV'18)                  & 48.3        & 75.6        & 84.5        & 35.7          & 63.6          & 74.1          & 63.6                & 56.1          & 86.3          & 92.9          & 44.6          & 78.8          & 89   & 74.6                \\ \hline
		DAN \cite{DAN_CVPR2017} (CVPR'17)                   & 41.4        & 73.5        & 82.5        & 31.8          & 61.7          & 72.5          & 60.6                & -             & -             & -             & -             & -             & -    & -                   \\
		JGCAR \cite{wang2018joint} (MM'18)  		& 44.9		& 75.3		& 82.7		& 35.2		& 62.0	& 72.4  & 62.1   & 52.7     & 82.6		& 90.5		& 40.2		& 74.8		& 85.7      & 71.1 \\
		LTBN (Emb) \cite{two_branch_TPAMI2018} (TPAMI'18)     & 43.2        & 71.6        & 79.8        & 31.7          & 61.3          & 72.4 & 60.0               & 54.9          & 84.0          & 92.2          & 43.3          & 76.4          & 87.5  & 73.1                \\
		RRF \cite{RRF-Net_ICCV2017} (ICCV'17)                   & 47.6        & 77.4        & 87.1        & 35.4          & 68.3          & 79.9          & 66.0                & 56.4          & 85.3          & 91.5          & 43.9          & 78.1          & 88.6  & 74.0                \\
		VSE++ \cite{vse++} (BMVC'18)                 & 52.9        & 79.1        & 87.2        & 39.6          & 69.6          & 79.5          & 68.0                & 64.6          & 89.1          & 95.7          & 52.0          & 83.1          & 92.0  & 79.4                \\
		GXN \cite{GXN_CVPR2018} (CVPR'18)                   & -           & -           & -           & -             & -             & -             & -                   & 68.5          & -             & \textbf{97.9}          & 56.6          & -             & 94.5 & -                   \\
		SCO \cite{SCO_CVPR2018} (CVPR'18)                   & 55.5        & 82.0        & 89.3        & 41.1          & 70.5          & 80.1          & 69.8                & 69.9          & 92.9          & 97.5          & 56.7          & 87.5          & 94.8 & 83.2                \\ 
		SCAN (T2I) \cite{SCAN_ECCV2018} (ECCV'18)             & 61.8        & 87.5        & 93.7        & 45.8          & 74.4          & 83.0          & 74.4                & 70.9          & 94.5          & 97.8          & 56.4          & 87.0          & 93.9 & 83.4                \\
		SCAN (I2T) \cite{SCAN_ECCV2018} (ECCV'18)             & \textbf{67.9}        & \textbf{89.0}        & \textbf{94.4}        & 43.9          & 74.2          & 82.8          & 75.4                & 69.2          & 93.2          & 97.5          & 54.4          & 86.0          & 93.6 & 82.3                \\ \hline
		\textbf{MTFN}              & 63.1        & 85.8        & 92.4       & 46.3 & 75.3          & 83.6          & 74.4                & 71.9          & 94.2          & \textbf{97.9}          & 57.3          & 88.6          & \textbf{95.0} & 84.2                \\
		\textbf{MTFN-RR w/o $S_{TT}\left(T, T'\right)$}			  &65.3			&88.3		  &93.3 		&46.7		  &75.9		 	&83.8			&75.6			&\textbf{74.3}		&\textbf{94.9}		&\textbf{97.9}			&57.5			&88.8			&\textbf{95.0}		&84.7						\\
		\textbf{MTFN-RR with $S_{TT}\left(T, T'\right)$}           & 65.3        & 88.3        & 93.3        & \textbf{52.0} & \textbf{80.1} & \textbf{86.1} & \textbf{77.5}       & \textbf{74.3} & \textbf{94.9} & \textbf{97.9} & \textbf{60.1} & \textbf{89.1} & \textbf{95.0} & \textbf{85.2}       \\ \hline \hline
	\end{tabular}
	\label{tab:results_overall}%
\end{table*}

\section{Experiment}
\subsection{Experimental Setup}
\noindent\textbf{Datasets and Evaluation Metric.}
We conducted several experiments on two widely used  datasets, \ies, Flickr30k \cite{flickr30k} and MSCOCO \cite{MSCOCO} with the following widely-used experimental protocols:
1) \textbf{Flickr30k} contains 31000 images collected from the Flickr webset. Each image is manually annotated by 5 sentences. We use the same data split setting as in \cite{vse++, SCAN_ECCV2018} with the training, validation and test splits containing 28000, 1000 and 1000 images, respectively.
2) \textbf{MSCOCO} consists of 123287 images and each one is associated with 5 sentences.
We use the public training, validation and testing splits following \cite{SCAN_ECCV2018, vse++}, where 113287 and 5000 images are used for training and validation, respectively. For the 5000 test images, we report results by i) averaging over 5 folds of 1k test images and ii) directly evaluating on the full 5k images.

We conduct two kinds of image-text matching tasks: 1) sentence retrieval, \ies, retrieving groundtruth sentences given a query image (I2T); 2) image retrieval, \ies, retrieving groundtruth images given a query text (T2I).
The commonly used evaluation metric for the I2T and T2I tasks is $R@L$ defined as the recall rate at the top $L$ results to the query, and usually $L = \{1, 5, 10\}$.
We also used ``mR" score proposed in \cite{SCO_CVPR2018} for additional evaluation, which averages all the recall scores of $R@L$ to assess the overall performance for both I2T and T2I tasks.

\noindent\textbf{Implementation Details.}
The feature extraction in our experiment generally follows the pre-process adopted in \cite{anderson_vqa, SCAN_ECCV2018}.
Specifically, for visual feature representation, we use the ResNet \cite{resnet} model to extract the CNN features for 36 regions detected by pre-trained Faster-RCNN \cite{faster_rcnn} model on Visual Genomes \cite{genomes}.
Then after global average pooling on the feature map, an image can be represented by a 2048d global feature vector.
For textual feature representation, we use a GRU \cite{gru} initialized with the parameters of a pre-trained Skip-thoughts model \cite{skip-thoughts} to represent each text sentence by a 2400d feature vector.
We trained our model using Adam optimizer with a mini-batch size of 128 for 50 epochs on each dataset.
The initial learning rate is 0.0001, decayed by 2 every 10 epochs.
The two fusion branches in our model are trained successively, where we use the parameters of the Image-Text fusion branch to initialize the Text-Text fusion branch for stable training performance.
The parameters $d_{v}$, $d_{t}$, $d_{f}$ are empirically set as 1024, the margin $\alpha$ is set to 0.2 and $R$ is 20.
In the cross-modal RR, the number $K$ for nearest-neighbor searching is respectively set to 15 and 7 for Flickr30k and MSCOCO datasets.
Our model is implemented in PyTorch \cite{pytorch} and all the experiments are conducted on a workstation with two NVIDIA 1080 Ti GPUs.

\subsection{Comparisons with the State-of-the-arts}
We compared our model with several recent state-of-the-art models, including the classification-based methods: LTBN (Sim) \cite{two_branch_TPAMI2018}, sm-LSTM \cite{sm-LSTM_CVPR2017}, CMPM \cite{CMPM_ECCV2018} and the embedding-based methods: DAN \cite{DAN_CVPR2017}, JGCAR \cite{wang2018joint}, LTBN (Emb) \cite{two_branch_TPAMI2018}, RRF \cite{RRF-Net_ICCV2017}, VSE++ \cite{vse++}, GXN \cite{GXN_CVPR2018}, SCO \cite{SCO_CVPR2018}, SCAN \cite{SCAN_ECCV2018}.
Note that for fair and objective comparison, feature extractions for images and text and evaluation protocols in all methods are consistent with  \cite{SCAN_ECCV2018,SCO_CVPR2018}.

\begin{figure*}[!htb]
	\centering
	\includegraphics[width=0.92\textwidth]{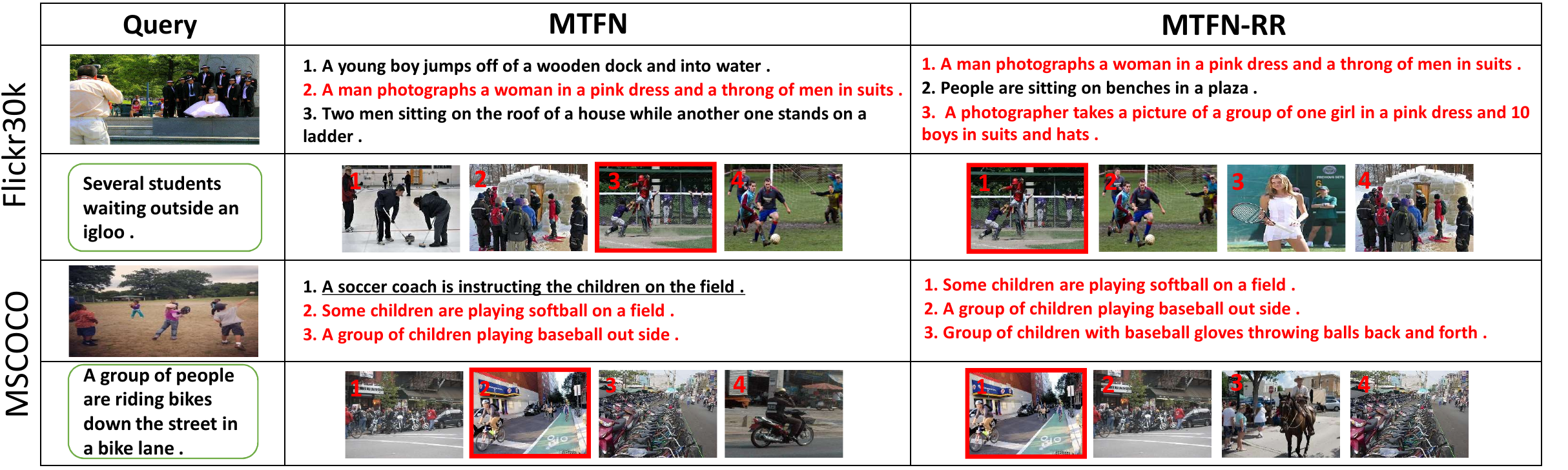}
	\caption{Quantitative results of I2T and T2I retrieval on Flickr30k and MSCOCO datasets obtained by our MTFN and MTFN-RR models. For I2T, the ground-truth text are marked as red and bold, while the text sharing similar meanings are marked with underline. For T2I, the groundtruth images are outlined in red rectangles. More results can be referred to our supplementary.}	
	\label{fig:example}
\end{figure*}

Table \ref{tab:results_overall} shows the overall I2T and T2I retrieval results of our model and the counterparts on the Flickr30k and MSCOCO datasets.
We can make the following observations:
\begin{itemize}
	\item Our MTFN itself achieves competitive results for both tasks on the two datasets.
	It indicates that our proposed fusion network is capable to learn the effective similarity function to fully encode the interactions between image and text.
	We note that our MTFN obtains slightly inferior I2T performance than the current best model SCAN on Flickr30k.
	However, the SCAN method still cannot outperform on both I2T and T2I task with one model.
	A probable reason is the smaller size of Flickr30k compared to MSCOCO. 
	However, the difference in performance between MTFN and SCAN is insignificant compared to immense difference in algorithmic complexity: SCAN is much more complex than our MTFN as it is elaborately designed for I2T and T2I tasks separately, and relies on fine-grained local features of both image regions and textual words with additional attention mechanism.
	\item When combining with the proposed cross-modal RR scheme with text-text similarity $S_{TT}$, our MTFN-RR gains remarkable improvements compared with MTFN on both tasks and achieves the state-of-the-art performance in most cases.
	The main reason is that the two cascaded steps of the framework exploit the synergy between the I2T and
	T2I retrieval tasks by looking at the image-text matching
	task simultaneously from two perspectives (from image to text and from text to image). 
	In addition, we also explore our MTFN-RR without exploiting text-text similarity.
	From the results we can see the improvement on T2I decreases significantly due to the data imbalance between images and text.
	\item The notable improvement of our MTFN-RR is achieved on the R@1 and R@5 on both datasets, which is more beneficial for retrieval in practice.
	Specifically, on Flickr30k dataset, the best R@1 on T2I task of our MTFN-RR is 52.0, which is superior to SCAN with a large margin of 13.5\%.
	On MSCOCO 1k test, our MTFN-RR obtains R@1 score 74.3 and 60.1 on I2T and T2I tasks, consistently outperforming the second best by 4.8\% and 6.0\%, respectively.
	Besides, our model also performs best on MSCOCO 5k test shown in Table \ref{tab:results_mscoco5k}, which further verifies the superiority of the proposed MTFN-RR.
	\item The improvement on the T2I task by our MTFN-RR is more remarkable than that on the I2T task, showing the advance of the Text-Text fusion in our proposed fusion network on capturing the similarity of semantically related text and enhancing the accuracy of the learned image-text similarity.
	Fig. \ref{fig:example} visualizes several typical retrieval examples obtained by our MTFN and MTFN-RR on the two datasets.
\end{itemize}


\begin{table}[!htb]
	\centering
	\footnotesize
	\caption{The I2T and T2I retrieval results obtained by our models and the counterparts on MSCOCO 5k
		test set.}
	\begin{tabular}{c|ccc|ccc}
		\hline \hline
		\multirow{2}{*}{Method} & \multicolumn{3}{c|}{I2T} & \multicolumn{3}{c}{T2I} \\
		\cline{2-7}          & R@1 & R@5 & R@10  & R@1 & R@5 & R@10 \\
		\hline
		DPC \cite{DPC_arxiv2017}  & 41.2  & 70.5  & 81.1  & 25.3  & 53.4  & 66.4  \\
		GXN \cite{GXN_CVPR2018}  & 42.0  & -     & 84.7  & 31.7  & -     & 74.6  \\
		SCO \cite{SCO_CVPR2018}  & 42.8  & 72.3  & 83.0  & 33.1  & 62.9  & 75.5  \\
		CMPM \cite{CMPM_ECCV2018} & 31.1  & 60.7  & 73.9  & 22.9  & 50.2  & 63.8  \\
		SCAN (T2I) \cite{SCAN_ECCV2018} & 46.2  & 77.1  & 86.8  & 34.3  & 64.7  & 75.8  \\
		SCAN (I2T) \cite{SCAN_ECCV2018} & 46.4  & 77.4  & 87.2  & 34.7  & 64.8  & \textbf{76.8}  \\
		\hline
		\textbf{MTFN (Ours)}  & 44.7  & 76.4  & \textbf{87.3}  & 33.1  & 64.7  & 76.1  \\
		\textbf{MTFN-RR (Ours)} & \textbf{48.3} & \textbf{77.6} & \textbf{87.3}  & \textbf{35.9} & \textbf{66.1} & 76.1  \\
		\hline \hline
	\end{tabular}%
	\label{tab:results_mscoco5k}%
\end{table}%

\begin{table*}[!htb]
	\centering
	\small
	\caption{Comparison of our MTFN with other common fusion strategies on the MSCOCO 1k test set. Check mark represents the combination of different fusion methods and attention mechanism.}
	\begin{tabular}{|cccc|c|c|c|ccc|ccc|}
		\hline
		\multicolumn{4}{|c|}{Fusion Strategy} & \multirow{2}{*}{Attention} & \multirow{2}{*}{\begin{tabular}[c]{@{}c@{}}Training\\ Time (hours)\end{tabular}} & \multirow{2}{*}{\begin{tabular}[c]{@{}c@{}}Evaluating\\ Time (s)\end{tabular}} & \multicolumn{3}{c|}{I2T} & \multicolumn{3}{c|}{T2I} \\ \cline{1-4} \cline{8-13} 
		\textbf{MTFN (Ours)} & Sum & Product & Concatenation &  &  &  & R@1 & R@5 & R@10 & R@1 & R@5 & R@10 \\ \hline
		& $\checkmark$ &  &  &  & 7.9 & 36.2 & 65.2 & 90.1 & 96.2 & 45.3 & 82.6 & 90.8 \\
		&  & $\checkmark$ &  &  & 7.9 & 36.7 & 67.1 & 92.8 & 97.2 & 48.3 & 84.6 & 92.5 \\
		&  &  & $\checkmark$ &  & 8.2 & 37.6 & 65.1 & 90.6 & 96.1 & 46.2 & 83.0 & 91.9 \\ \hline
		& $\checkmark$ &  &  & $\checkmark$ & 47.1 & 261.5 & 66.3 & 90.2 & 96.2 & 46.1 & 82.9 & 90.7 \\
		&  & $\checkmark$ &  & $\checkmark$ & 48.1 & 262.9 & 67.3 & 92.6 & 97.1 & 48.8 & 84.8 & 92.6 \\
		&  &  & $\checkmark$ & $\checkmark$ & 48.9 & 264.3 & 66.1 & 91.8 & 96.4 & 46.6 & 83.2 & 92.0 \\ \hline
		{$\checkmark$} &       &       &       & {$\checkmark$} & 50.2  & 283.2    & 70.8  & 92.8  & 97.1  & 53.7  & 87.2  & 94.5  \\
		$\checkmark$ &  &  &  &  & 9.0 & 40.2 & \textbf{71.9} & \textbf{94.2} & \textbf{97.9} & \textbf{57.3} & \textbf{88.6} & \textbf{95.0} \\ \hline
	\end{tabular}
	\label{tab:result_fusion}%
\end{table*}%

\begin{figure*}[!htb]
	\centering
	\includegraphics[width=0.85\textwidth]{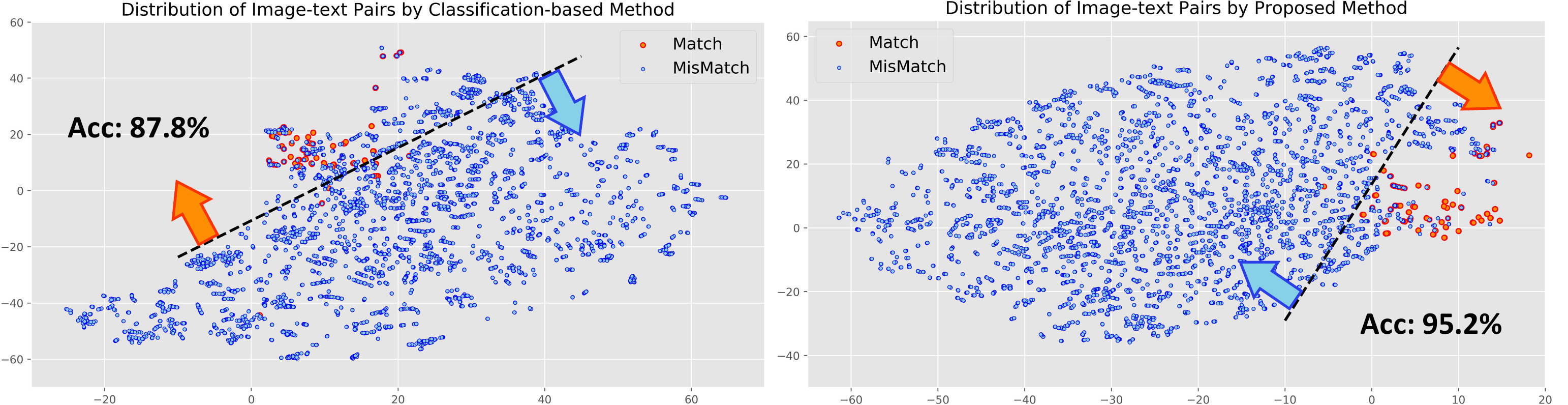}
	\caption{Visualization of the fusion vector $\mathbf{f}$ by classification-based method and our MTFN embedding on the part of MSCOCO test set (8000 image-text pairs) with the learned linear SVM boundary.}	
	\label{fig:tsne}
\end{figure*}

\subsection{Further Analysis}
\noindent\textbf{Distribution of Fusion Vector.}
We concluded our qualitative analysis by providing a global view of the performance of our proposed MTFN comparing to the classification-based method by replacing ranking loss in MTFN with the logistic regression.
We visualize the distributions of fusion vector $\mathbf{f}$ on MSCOCO dataset by deploying the t-SNE tool to map it onto a 2D space.
Additionally for better analysis, we further use a standard SVM (support-vector machine) for each embedding to find a linear boundary between ``match'' and ``mismatch'' dots and to compute classification accuracy.
Fig. \ref{fig:tsne} depicts the overall results and the black dashed line denotes the learned SVM boundary.
We can conclude that our model can better preserve the structure of matching image-text pairs with a larger margin and get much higher accuracy for classifying ``match'' and ``mismatch''.
The main reason is that our ranking-based loss optimizes the model in terms of a margin without forcing the image-text pairs only to ``1' and ``-1''.

\noindent\textbf{Analysis on Fusion Strategy.}
In this experiment, we compare our MTFN with previous linear fusion schemes used in \cite{RRF-Net_ICCV2017, two_branch_TPAMI2018}, \egs, element-wise sum/product and concatenation, by evaluating the I2T and T2I retrieval results and the training and evaluating time consumption for time complexity.
Besides, the popular attention mechanism used in \cite{attention, DAN_CVPR2017, SCAN_ECCV2018}
is also included to assess its influence on different fusion schemes.
Following the experimental setting in \cite{MCB_EMNLP2016}, each combination of model has similar amount of model parameters by combining with multiple fully connected layers.

Table \ref{tab:result_fusion} shows that our MTFN itself outperforms all the traditional linear fusion strategies with much less training and evaluating time.
The attention mechanism has beneficial impact on the previous linear fusion schemes, however, it deteriorates the performance of our MTFN and greatly increases the time consumed.
The potential reason is that our MTFN already effectively encodes the bilinear interactions between the global features of images and text with rank constraint.
Using attention mechanism just leads to a great increase of model complexity and time consumption.

\begin{figure}[!htb]
	\centering
	\includegraphics[width=0.39\textwidth]{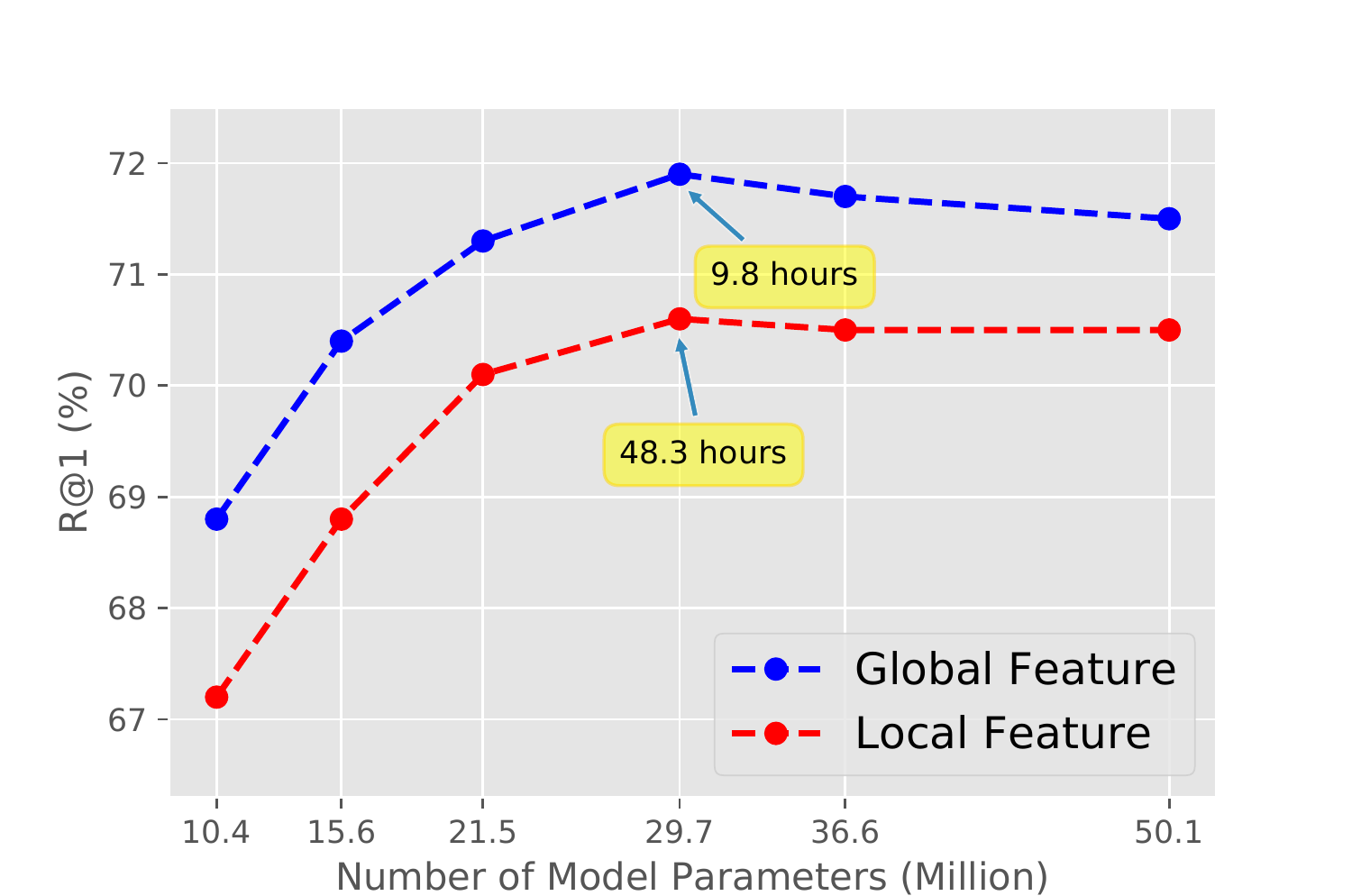}
	\caption{R@1 scores of the I2T retrieval on the MSCOCO 1k test set with various model parameters using global and local features. The yellow label indicates the time consumption when achieving best result.}	
	\label{fig:example_abla2}
\end{figure}

\noindent\textbf{Analysis on Model Complexity.}
Our MTFN is flexible to use either global or local features for images and text on tensor fusion constraint $R$ as depicted in Eq. \ref{elementwise_r}.
To investigate the effect of raw features and hyper-parameter $R$ on our MTFN, we take the MSCOCO dataset as testbed to assess the model complexity using the extracted global or local features for images and text.
As shown by the comparison results in Fig. \ref{fig:example_abla2}, we can observe that with various quantities of model parameters, using global features can consistently obtain better performance than using local ones, showing that the bilinear tensor fusion process in our MTFN is more effective to handle the global features.
Moreover, to achieve the best performance, our MTFN can be trained much faster (around \textbf{9} hours) using global features than the time (around 48 hours) using local ones.
It is worthy mentioning that we also evaluate the model complexity of previous sm-LSTM and SCAN methods.
In practice, they need around \textbf{50} and \textbf{60} hours for training, respectively.
Thus, it further demonstrates that our MTFN is much more efficient than these two counterparts using local features for training, due to the superiority of tensor fusion applied in our MTFN.
\begin{figure}[!htb]
	\centering
	\subfigure[]{\label{fig:eval_reranking}
		\centering
		\includegraphics[width=0.243\textwidth]{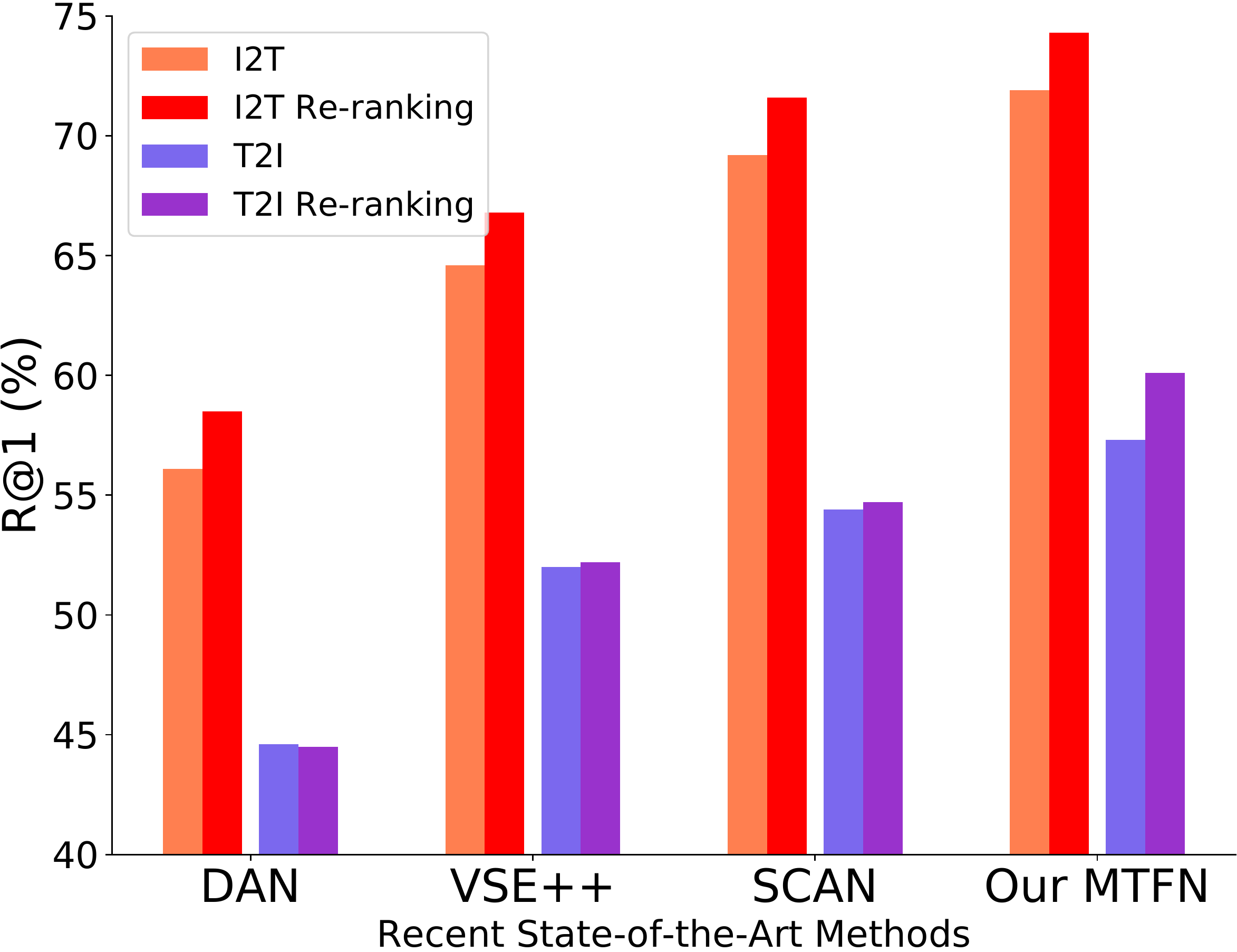}
	}
	\hspace{-.12in}
	\subfigure[]{\label{fig:re-rank_k}
		\centering
		\includegraphics[width=0.22\textwidth]{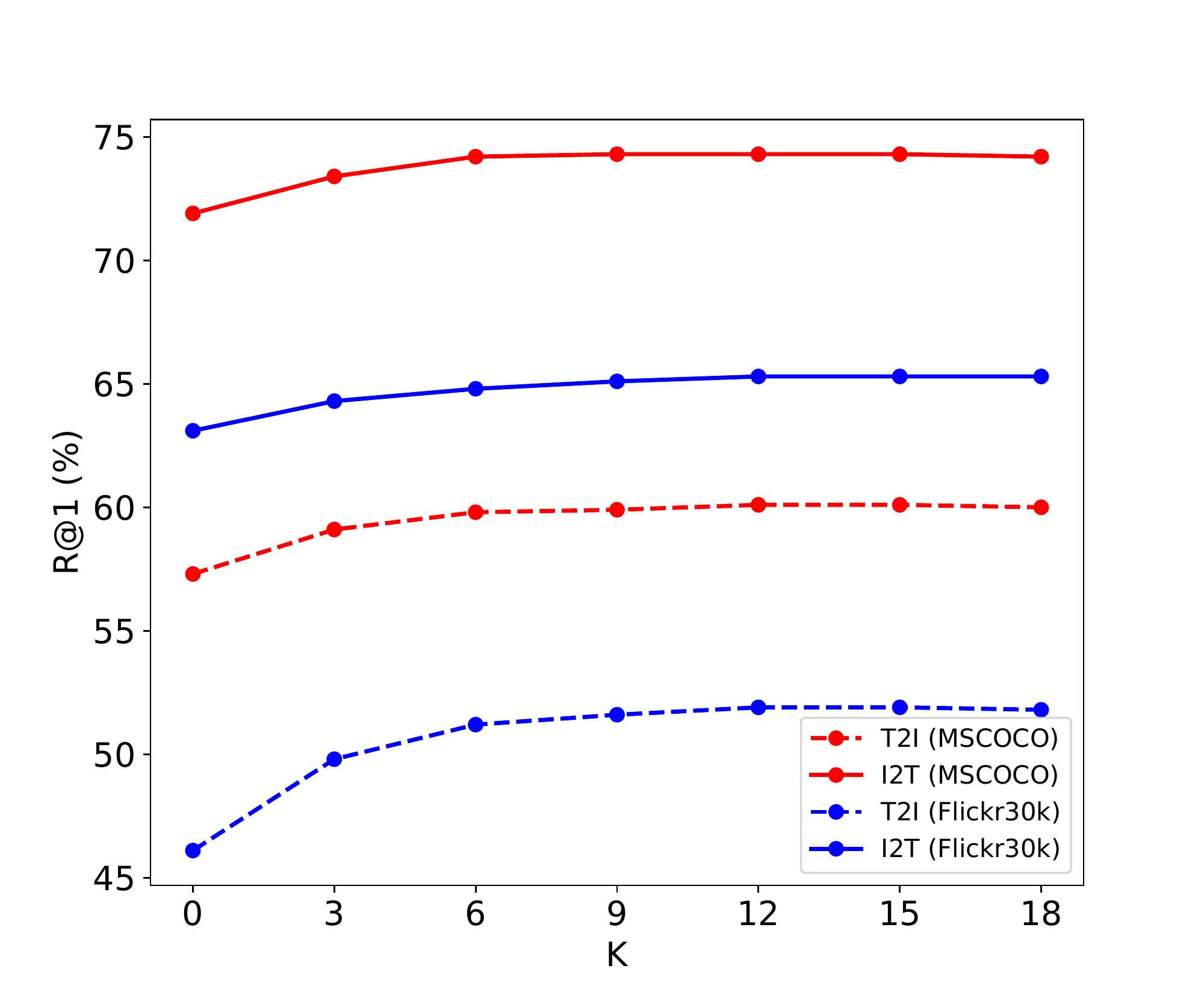}
	}
	\vspace{-0.1in}
	\caption{For the proposed RR scheme: (a) Comparison of RR applied to our MTFN and resent state-of-the-art methods on MSCOCO dataset. (b) Effect of the nearest-neighbor $K$ used in RR on R@1 on Flickr30k and MSCOCO datasets.}	
\end{figure}

\noindent\textbf{Analysis on Cross-modal Re-ranking.}
As aforementioned, the proposed cross-modal RR scheme can also be applied to \textit{most previous methods} that utilize image-text similarity to obtain a ranking list.
In this experiment, we take MSCOCO dataset and apply our proposed RR scheme on our MTFN and three latest methods DAN \cite{DAN_CVPR2017}, VSE++ \cite{vse++} and SCAN \cite{SCAN_ECCV2018}, to investigate its effect on refining the retrieval results.
Specifically, for each query instance (image or sentence), we perform I2T and T2I for each model to  get its initial retrieval ranking list.
Then we can obtain its refined ranking list after re-ranking process.
Fig. \ref{fig:eval_reranking} shows the results in terms of $R@1$ on both I2T and T2I tasks by comparing the initial and refined ranking lists.
It is obvious that the re-ranking process makes remarkable improvements for all the four methods on the I2T task, showing that utilizing the cross-modal associations helps in achieving more accurate retrieval.
Besides, we can also observe that re-ranking is effective for our MTFN on the T2I task while its effect is minor in other cases.
In Fig. \ref{fig:re-rank_k}, we also assess the impact of various nearest neighbors $K$ on our MTFN with R@1 during the RR process.
We can see that the RR performance on R@1 remains stable for large neighborhood, with the critical point at $K$=6, below which the performance degrades. 


\noindent\textbf{Analysis on Model Ensemble.}
Model ensemble is a practical strategy that integrates the retrieval results from multiple models.
The latest approaches of RRF-Net and SCAN have already studied the effect of model ensemble and show its effectiveness to further boost the retrieval performance. 
In this part we follow them to integrate the strength of averaging $M$ individual MTFN-RR model and compare to RRF-Net and SCAN with different cases of model ensemble on Flickr30k dataset.
Specifically, for RRF-Net and our MTFN-RR, $M$ denotes the number of individual model used for ensemble, while (I2T + T2I) denotes the integration of the SCAN models separately trained for I2T and T2I.
As the result shown in Table \ref{tab:ensemble}, for our MTFN-RR model, compared with a single model (\ies, $M = 1$), merging multiple models $M = 2, 3$ generally obtains much better retrieval performance without increasing the training complexity.
In addition, our MTFN-RR ($M = 3$) significantly outperforms the best ensemble result of SCAN (I2T + T2I) on T2I task, showing the advantage of our MTFN-RR method.

\section{Conclusion}
\label{sec:conc}
In this work, we proposed a novel image-text matching method named MTFN, which directly learns the image-text similarity function by multi-modal tensor fusion of global visual and textual features effectively, without redundant training steps.
We then combined our MTFN with an effective and general cross-modal RR scheme, \ies, the MTFN-RR framework, to boost the I2T and T2I retrieval results considering additional unimodal text-text similarity.
Experiments on two benchmark datasets showed the effectiveness of our MTFN and the RR scheme, which achieve the state-of-the-art retrieval performance with much less time consumption. 
In the future, we consider to develop more effective cross-modal RR schemes to form an end-to-end matching framework.

\begin{table}[!htb]
	\centering
	\small
	\caption{Model ensemble results of our MTFN-RR and the counterparts RRF-Net and SCAN on Flickr30k dataset.}
	\begin{tabular}{|c|c|c|c|c|}
		\hline
		\multirow{2}{*}{Ensemble Model} & \multicolumn{2}{c|}{I2T} & \multicolumn{2}{c|}{T2I} \\ \cline{2-5}
		& R@1                     & R@5           & R@1               & R@5              \\ \hline
		RRF-Net ($M = 3$) \cite{RRF-Net_ICCV2017}                   & 50.3                    & 79.2          & 37.4 & 70.4             \\
		SCAN (I2T + T2I) \cite{SCAN_ECCV2018}                  & 67.4                    & \textbf{90.3}          & 48.6              & 77.7             \\
		MTFN-RR $M = 1$        & 65.3                    & 88.3          & 52.0              & 80.1            \\
		MTFN-RR $M = 2$        & 67.4                    & 89.4          & 52.8 & 80.9             \\
		MTFN-RR $M = 3$        & \textbf{67.7}           & 90.1          & \textbf{53.2}     & \textbf{81.2} \\ \hline
	\end{tabular}
	\label{tab:ensemble}
\end{table}

\section{Acknowledgment}
This work was supported in part by the National Natural Science Foundation of China under grants No. 61602089, 61572108, 61632007 and the Sichuan Science and Technology Program, China, under Grants No. 2019ZDZX0008 and 2018GZDZX0032.

%
\bibliographystyle{ACM-Reference-Format}
\bibliography{MM2019-tan}

%

\end{document}